\documentclass[10pt,twocolumn,letterpaper]{article}

\usepackage[pagenumbers]{cvpr} % Public arXiv version with page numbers

\usepackage{amsmath}
\usepackage{amssymb}
\usepackage{booktabs}
\usepackage{placeins}
\usepackage{xspace}
\usepackage{CJKutf8}

\newcommand{\paper}{TextVBench\xspace}

\DeclareMathOperator{\concat}{concat}

\definecolor{cvprblue}{rgb}{0.21,0.49,0.74}
\usepackage[pagebackref,breaklinks,colorlinks,allcolors=cvprblue]{hyperref}

\def\paperID{*****} % *** Enter the Paper ID here
\def\confName{CVPR}
\def\confYear{2027}

\title{TextVBench: Evaluating Text Rendering and Preservation in Video Generation}

\author{
Ruiqiang Zhang$^{1,*}$ \quad Hengyi Wang$^{1,*}$ \quad Chang Liu$^{1}$ \\
Guanjie Wang$^{1}$ \quad Zehua Ma$^{1,\dagger}$ \quad Weiming Zhang$^{1,\dagger}$ \\
\textsuperscript{1}Anhui Province Key Laboratory of Digital Security,\\[-1pt]
University of Science and Technology of China \\
{\small\texttt{zrq\_ustc@mail.ustc.edu.cn} \quad
\texttt{why0\_1@mail.ustc.edu.cn}} \\
{\small\texttt{hichangliu@mail.ustc.edu.cn} \quad
\texttt{wangguanjie@mail.ustc.edu.cn}} \\
{\small\texttt{mzh045@ustc.edu.cn} \quad
\texttt{zhangwm@ustc.edu.cn}} \\
{\small $^{*}$Equal contribution \quad $^{\dagger}$Corresponding authors}
}
\hypersetup{
  pdftitle={TextVBench: Evaluating Text Rendering and Preservation in Video Generation},
  pdfauthor={Ruiqiang Zhang, Hengyi Wang, Chang Liu, Guanjie Wang, Zehua Ma, Weiming Zhang}
}

\begin{document}
\maketitle
\begin{abstract}
Modern video generators can synthesize convincing scenes while rendering a
few characters incorrectly, omitting requested text, or changing it across
frames. Measuring these failures is harder than applying OCR independently
to each frame: when a scene contains multiple text targets and multiple noisy
OCR regions, the evaluator must first determine which observation belongs to
which intended text. We introduce \paper, a benchmark and evaluation
framework that formulates visual-text evaluation as a multi-target,
multi-instance spatiotemporal correspondence problem. \paper contains two
complementary tracks: TextVBench-T2V measures the creation and maintenance of
prompt-specified text, while TextVBench-Ref2V measures preservation of text
from reference images. Both tracks use target-conditioned candidate
selection rather than unconstrained VLM transcription, support one target
mapping to multiple OCR regions, and separate frame-level spatial fidelity
from cross-frame temporal fidelity. A length-weighted harmonic score
summarizes these factors, while extra-text diagnostics expose failure modes
hidden by a single OCR score. The benchmark is organized around
controlled text, layout, carrier, and motion attributes so that aggregate
scores can be traced back to specific capability limits.
Across the evaluated model runs, MiniMax-H3 reaches a T2V TextVScore of
68.32, while the strongest Ref2V score is 56.51; attribute slices reveal a pronounced drop as the number
of simultaneous reference text regions increases.
\end{abstract}
    
\section{Introduction}
\label{sec:intro}

Video generation has advanced from short, unstable clips to visually
convincing scenes with coherent subjects, camera motion, and object dynamics.
Yet text remains a revealing failure mode. A generated advertisement can look
plausible while misspelling a brand, a phone screen can silently replace a
number, and a correct word can flicker into a different one for a single frame.
Such errors are not minor perceptual artifacts: a one-character change can
alter the meaning of the generated content.

The difficulty is not limited to generation. Reliable evaluation itself is
non-trivial once a prompt or reference contains several text targets. OCR can
split one semantic phrase across multiple boxes, merge adjacent targets, or
detect repeated and hallucinated text. Consequently, an evaluator does not
receive a ready-made pair of prediction and ground truth. It must first answer
a more basic question: \emph{which observed region or group of regions
corresponds to each intended text target?}

Most image-domain visual-text evaluations avoid this ambiguity through a
single word, a short string, or a recoverable reading order
\cite{tuo2023anytext,tai2025textcrafter,fallah2025textinvision}. Global string
comparison can fail when two correctly rendered targets occupy different
semantic regions, while one-to-one set assignment cannot represent a target
split across OCR boxes. These are representation failures before they are
recognition failures: assignment is only meaningful after observations have
been organized into semantic units.

Video adds a second axis of ambiguity. Text may be temporarily occluded,
leave the camera view with its carrier, reappear after a viewpoint change, or
become corrupted for only part of a clip. High mean frame accuracy therefore
does not imply stable text. More fundamentally, video turns static matching
into spatiotemporal state tracking: the target, its carrier, the current visual
observation, and its identity over time must remain connected.

We introduce \paper to evaluate this structured problem. TextVBench-T2V
tests whether a text-to-video model creates prompt-specified text on the
intended carrier and maintains it over time. TextVBench-Ref2V tests whether an
image-to-video model preserves visual text extracted from one or more reference
images. The two tracks share a target-centric evaluator that first obtains OCR
candidates, then establishes semantic correspondence using explicit candidate
IDs and marker images, and only then computes spatial and temporal fidelity.
The vision-language model (VLM), Qwen3.8-27B in our evaluation, selects among
OCR candidates; it is not allowed to freely transcribe or repair their text.

\begin{figure*}[t]
  \centering
  \includegraphics[width=\textwidth]{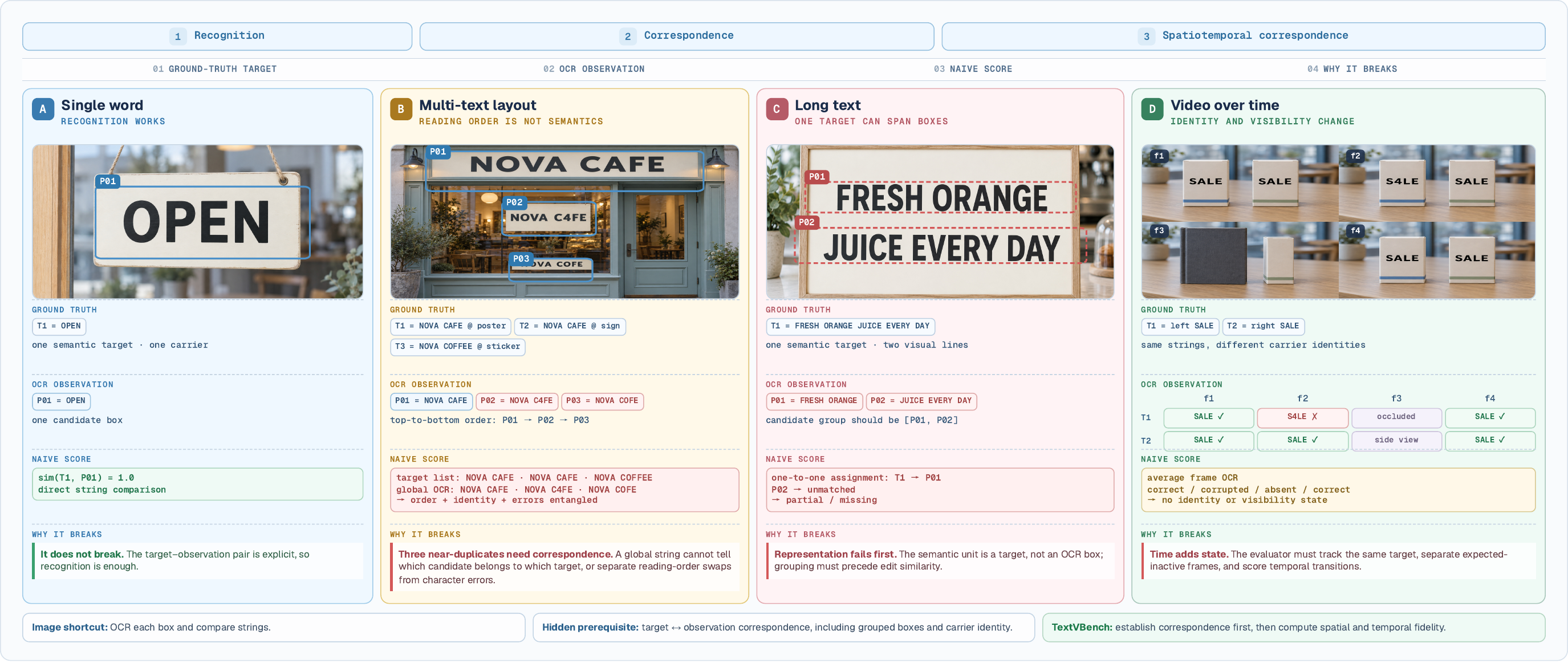}
  \caption{Why visual-text evaluation is difficult. The aligned cases progress
  from single-target recognition to multi-target correspondence, fragmented
  OCR observations, and identity-aware tracking over time. A reliable
  evaluator must associate OCR candidates with semantic targets before scoring
  spatial and temporal fidelity.}
  \label{fig:motivation}
\end{figure*}

Our contributions are threefold:
\begin{itemize}
  \item We formulate visual-text evaluation in generated video as a
  multi-target, multi-instance spatiotemporal correspondence problem, exposing
  assumptions that make image-domain OCR scores unreliable in structured
  scenes.
  \item We introduce complementary T2V and Ref2V tracks whose controlled
  attribute spaces cover text creation and reference preservation under
  layout, carrier, motion, and viewpoint variation.
  \item We develop a target-centric automatic evaluator that supports grouped
  OCR observations, carrier/reference routing, spatial and temporal fidelity,
  and auditable diagnostics for per-target missing evidence and extra text from
  repeated or hallucinated observations.
\end{itemize}

\section{Related Work}
\label{sec:related}

\paragraph{Visual text generation.}
Image generation methods increasingly use glyph, layout, or OCR supervision to
improve rendered text. GlyphControl and TextDiffuser-2 make this control
explicit through glyph-conditioned or language-model-assisted rendering
\cite{yang2023glyphcontrol,chen2024textdiffuser2}. AnyText introduced multilingual generation and editing
with an OCR-aware embedding pathway and a dedicated benchmark
\cite{tuo2023anytext}; subsequent benchmarks broadened evaluation toward prompt
and text complexity \cite{fallah2025textinvision}. TextCrafter introduces
CVTG-2K for multiple text instances, lengths, positions, and attributes, and
evaluates long-string rendering on LongText-Bench
\cite{tai2025textcrafter}. These efforts establish that visual text deserves
specialized evaluation, but their static setting commonly assumes a simple
instance or an order from which a prediction--target alignment can be
recovered. \paper focuses on the correspondence that precedes scoring when
several semantic targets, fragmented OCR observations, carriers, and times are
all present.

\paragraph{Video generation evaluation.}
General-purpose suites evaluate appearance, prompt alignment, motion, and
temporal quality. Reference-free semantic metrics such as CLIPScore show the
value of learned image--text alignment, but do not expose character-level
omissions or target identity \cite{hessel2021clipscore}. VBench decomposes video
quality into interpretable
dimensions \cite{huang2023vbench}, VBench++ extends this protocol to broader
T2V and I2V settings \cite{huang2024vbenchpp}, and EvalCrafter combines a wide
prompt set with multiple objective metrics and human-aligned aggregation
\cite{liu2023evalcrafter}. AVGen-Bench and MSAVBench broaden evaluation to
task-driven audio-video generation and multi-shot narratives, respectively
\cite{zhou2026avgenbench,wei2026msavbench}. These benchmarks are valuable for
overall generative quality, but their visual-text component does not establish
target-level spatial assignment or preserve text identity through time.

\paragraph{Visual text in video.}
T2VTextBench directly demonstrates that contemporary video generators struggle
with on-screen text fidelity and temporal consistency, using human evaluation
across diverse systems \cite{guo2025t2vtextbench}. Human judgment is flexible
and semantically aware, but it is expensive to repeat at training or
leaderboard scale. \paper complements this direction with an automatic,
auditable protocol for multi-target correspondence. Video text spotting systems
also model text identity across time rather than recognizing each frame in
isolation \cite{wu2024transdetr}; however, they generally assume transcription
of naturally occurring video text, whereas our task must first map generated
observations to prompt- or reference-defined targets. \paper's outputs retain the
selected OCR candidate IDs and per-target trajectories, enabling both aggregate
comparison and inspection of individual decisions.

\section{Why Visual-Text Evaluation Is Hard}
\label{sec:challenges}

\subsection{Image-Domain Limitations}

\paragraph{Correspondence is usually hidden by simplification.}
For a single sign containing one requested word, the target and OCR output are
naturally paired. Realistic scenes are less convenient. A long target such as
FRESH ORANGE JUICE EVERY DAY may span several lines and yield several OCR
boxes. Conversely, one OCR box may merge adjacent targets. Comparing each box
with one target penalizes a correct rendering because one semantic unit need not
equal one detector region.

\paragraph{Reading order is not semantic order.}
Consider a prompt that requests B in the upper-left and A in the lower-right.
The rendering may satisfy both spatial constraints while a reading-order
concatenation returns BA and a target list happens to store AB. A global edit
distance then reports an error that the image does not contain. The evaluator
must associate content with the specified region or carrier before comparing
strings.

\paragraph{One-to-one assignment is insufficient.}
Modern scene-text detectors expose character- or region-level geometry rather
than guaranteed semantic instances \cite{baek2019craft,liao2020db}. Hungarian
assignment \cite{kuhn1955hungarian} improves over concatenation when
targets and predictions are already meaningful units. It does not decide
which OCR boxes should first be grouped, cannot naturally express a merged box
serving several targets, and ignores whether correct text is attached to the
wrong carrier. Matching one duplicate also leaves the semantic status of the
remaining copies unresolved. The problem is thus not assignment alone, but
the representation on which assignment operates.

\subsection{Video-Specific Challenges}

\paragraph{Accuracy and stability are different.}
A target rendered correctly in most frames but corrupted for a short interval
can retain a high frame-average score while visibly flickering. Spatial
fidelity asks whether the text is correct at an observation time; temporal
fidelity asks whether consecutive observations of the same target remain
consistent. Neither quantity substitutes for the other.

\vspace{-0.8em}
\paragraph{Absence is not always an error.}
Text can disappear because its carrier is occluded or leaves the view, not
because the model failed to render it. Viewpoint changes are especially
important in Ref2V: text on the front and side of a package is not
simultaneously observable. Evaluation therefore needs an explicit activation
decision at the carrier or reference level instead of treating every absent
OCR result as a miss.

\paragraph{Identity must persist.}
Repeated strings make tracking ambiguous: if two signs both read SALE and one
disappears, string equality alone cannot identify the surviving instance.
Carrier ownership, intended position, and reference identity provide the
structure needed to maintain a target--observation relation. These cases
motivate three design requirements used below: target-centric correspondence,
spatial fidelity conditioned on an active carrier/reference, and temporal
fidelity over observations of the same target.

\section{TextVBench: Setup and Overview}
\label{sec:overview}

\subsection{Problem Formulation}

Let a generated video be sampled at times $f\in\mathcal{F}$. Its structured
specification contains targets $\mathcal{T}=\{t_i\}_{i=1}^{N}$, where target
$t$ has an identifier, normalized string $y_t$, and a carrier or reference
identity. OCR produces frame-level candidates
$\mathcal{C}_f=\{c_{fj}\}_{j=1}^{M_f}$, each with an identifier, recognized
string, confidence, and polygon. This detector--recognizer interface follows
the conventional OCR decomposition into text-region localization and reading
\cite{smith2007tesseract,baek2019craft,liao2020db}. An OCR candidate is a
perceptual observation, not yet a semantic text instance.

The evaluator estimates a correspondence
$A_{tf}\subseteq\mathcal{C}_f$. This set may contain one candidate, several
candidates ordered for concatenation, or be empty. Let $a_{tf}\in\{0,1\}$
indicate whether the target's carrier/reference is active in frame $f$. When
$a_{tf}=1$, the prediction is
\begin{equation}
  \hat y_{tf}=\concat\left(\{\operatorname{text}(c):c\in A_{tf}\}\right),
  \label{eq:prediction}
\end{equation}
and an empty set represents a missing target. When $a_{tf}=0$, the frame is
excluded from that target's fidelity sequence rather than counted as correct
or missing.

Thus the evaluation unit is not an OCR box but a semantic target connected to
zero or more observations over time. We refer to recovering these relations as
multi-target, multi-instance spatiotemporal correspondence.

We normalize strings by removing whitespace and punctuation while preserving
case, then use character-level normalized edit similarity
\begin{equation}
  \operatorname{sim}(u,v)=1-\frac{d_{\mathrm{lev}}(u,v)}
  {\max(|u|,|v|)},
  \label{eq:ned}
\end{equation}
where $d_{\mathrm{lev}}$ is Levenshtein distance
\cite{levenshtein1966binary}. Correspondence is a prerequisite to this score:
Eq.~\ref{eq:ned} is evaluated only after the relevant candidate group has been
selected.

Unresolved correspondence and reference routing are decided by Qwen3.8-27B,
using temperature $0$, top-$p=0.8$, and a 512-token response cap. The model
returns candidate or reference IDs rather than free-form transcriptions; all
scored strings come from OCR outputs.

\subsection{Two Complementary Tracks}

\begin{figure*}[t]
  \centering
  \includegraphics[width=0.8\textwidth]{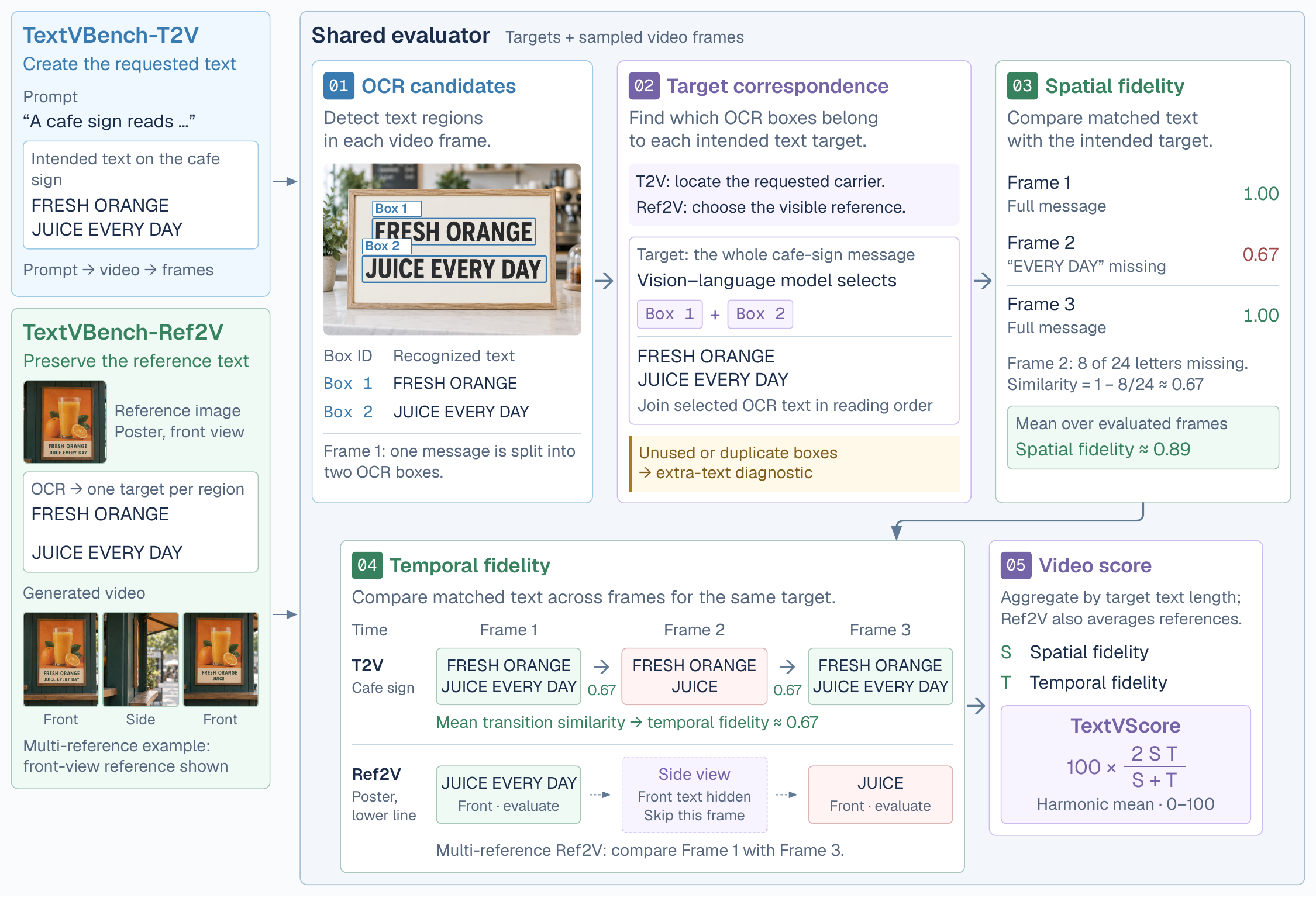}
  \caption{TextVBench overview and shared evaluator. A target is intended text
  from a T2V prompt or a Ref2V reference region; numbered boxes are OCR
  observations. In the T2V example, one target spans Box~1 and Box~2, whose
  text is joined without rewriting. Spatial fidelity compares observations
  with the target; temporal fidelity compares successive observations of the
  same target. Omitting ``EVERY DAY'' removes 8 of 24 normalized characters,
  giving the illustrated frame and transition similarities. Multi-reference
  Ref2V skips inactive views and compares text on reappearance. Length-weighted
  target aggregation (followed by equal reference averaging for Ref2V) yields
  $S$ and $T$ for TextVScore; extra text is reported as a separate diagnostic.
  Scenes, trajectories, and numerical examples are illustrative.}
  \label{fig:overview}
\end{figure*}

TextVBench-T2V starts from a text-to-video prompt and a structured set of text
slots. It evaluates whether the requested strings are created on their
intended carriers and maintained across the generated clip. TextVBench-Ref2V
starts from one or more reference images. Runtime reference OCR defines its
text slots, and evaluation asks whether those visual-text identities are
preserved in the generated video. As illustrated in
Fig.~\ref{fig:overview}, both tracks share OCR perception and target-conditioned
correspondence (steps 01--02). The resulting observations support two
complementary computations: target-relative spatial fidelity and cross-frame
temporal fidelity (steps 03--04), followed by video-level aggregation (step 05).
The source of the targets and the activation rule differ; the principle that
association must precede scoring does not. The numbered marker images follow
the visual-grounding strategy of presenting explicit region identifiers to a
multimodal model \cite{yang2023som}, while retaining the OCR strings as fixed
evidence.

\section{TextVBench-T2V}
\label{sec:t2v}

\subsection{Structured Construction}

\begin{figure*}[t]
  \centering
  \includegraphics[width=\textwidth]{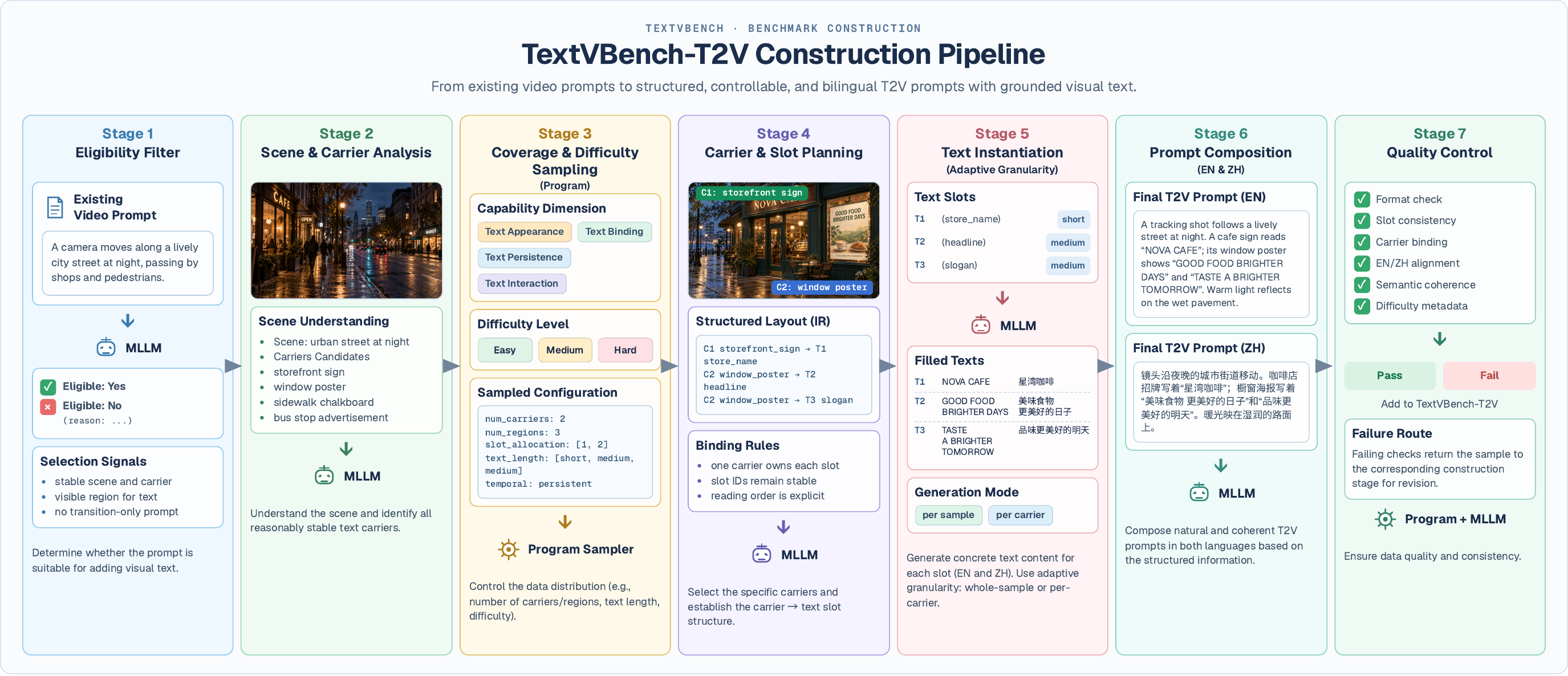}
  \caption{TextVBench-T2V construction pipeline. Seven stages turn source
  video prompts into controlled bilingual specifications: eligibility
  filtering, scene and carrier analysis, coverage and difficulty sampling,
  carrier--slot planning, text instantiation, prompt composition, and quality
  control. The example binds three text slots to a storefront sign and a
  window poster. Scene images are synthetic illustrations.}
  \label{fig:t2v-construction}
\end{figure*}

T2V examples are built from structured specifications rather than independent
free-form prompts. This defines a controlled generation difficulty space in
which each specification composes four groups of attributes:
(i) text attributes such as language and character/word length;
(ii) structural attributes such as target count and slot arrangement;
(iii) scene and carrier attributes, including signs, packages, screens, and
posters; and (iv) dynamic attributes describing camera, carrier, scale, and
viewpoint changes. This construction serves coverage, controllability, and
diagnosability: the same fields used to create examples become buckets for the
analysis in Sec.~\ref{sec:experiments}.

As shown in Fig.~\ref{fig:t2v-construction}, the implemented construction
pipeline first filters source scene descriptions
for natural visual-text insertion and proposes carrier candidates already
compatible with each scene. It then samples feasible easy, medium, and hard
variants. The current configuration uses one carrier and one--two slots for
easy examples, two carriers and two--three slots for medium examples, and
three carriers and three--five slots for hard examples. Text lengths progress
from short to medium/long accordingly. A structured scaffold binds every slot
to a selected carrier before bilingual English and Chinese strings and
parallel prompts are instantiated. Rule and semantic quality control operate
on the same specification, so neither language is produced by translating a
finished prompt from the other.

The main evaluation protocol considers persistent target strings: the carrier
and viewpoint may change, but the intended string does not. State-changing,
scrolling, or progressively written text requires a time-indexed ground-truth
sequence and falls within the TextVBench-Challenge scope discussed in
Sec.~\ref{sec:limitations}.

\subsection{Target Correspondence}

At each sampled frame, PaddleOCR \cite{cui2025paddleocr} supplies candidate
texts, confidences, and polygons. This detector--recognizer interface follows
the standard scene-text pipeline, in which region geometry and recognized text
are separate outputs \cite{smith2007tesseract,baek2019craft,liao2020db}.
Low-confidence candidates are removed;
exact duplicates beyond the multiplicity required by the targets and
candidates beyond the configured cap are retained as extra-text evidence.
The remaining candidates are drawn as colored polygons with explicit IDs in a
marker image, as shown in Fig.~\ref{fig:overview} (step 01).

Correspondence proceeds in two steps. First, a carrier-consistency judge
decides whether each required carrier is present and which candidate IDs
belong to it. This acts as a hard spatial boundary: text copied into a subtitle
cannot satisfy a target assigned to a poster. Second, candidates within the
carrier are associated with targets. A programmatic fast path accepts the
highest-similarity candidate when it reaches the configured threshold;
unresolved targets are sent in small
batches to a VLM together with the marker image and compact candidate list.
The VLM returns only selected IDs. It may select several IDs in visual reading
order for a fragmented target, but may not freely OCR, translate, or correct
candidate strings. Candidate IDs consumed by one request batch are removed
before the next, while reuse inside a batch permits a single merged OCR region
to support adjacent targets. Figure~\ref{fig:overview} (step 02) illustrates
one target selecting two ordered OCR regions before scoring.

\subsection{Spatial Fidelity}

For every carrier-present observation, spatial fidelity is
\begin{equation}
  s_{tf}=\begin{cases}
    \operatorname{sim}(y_t,\hat y_{tf}), & A_{tf}\neq\emptyset,\\
    0, & A_{tf}=\emptyset.
  \end{cases}
  \label{eq:spatial-frame}
\end{equation}
The target score $S_t$ is the mean over frames with $a_{tf}=1$. If its carrier
is never observed, $S_t=0$; this prevents a model from avoiding a text target
by omitting the entire carrier.

Unassigned active candidates, filtered duplicates, and capped candidates form
an extra pool. We report
\begin{equation}
  E_f=\frac{L(\mathcal{C}^{\mathrm{extra}}_f)}
  {L(\mathcal{C}^{\mathrm{extra}}_f)+L(\mathcal{C}^{\mathrm{matched}}_f)},
  \label{eq:extra}
\end{equation}
where $L$ counts normalized characters. ExtraTextRatio diagnoses repetition
and hallucination but is not folded into the main score.

\subsection{Temporal Fidelity}

The matched observations retain the same target identity across frames,
as illustrated in Fig.~\ref{fig:overview} (step 04).
For target $t$, sort its carrier-present observations by timestamp. For each
adjacent pair, define
\begin{equation}
  r_{t,k}=\begin{cases}
  \operatorname{sim}(\hat y_{t,k},\hat y_{t,k+1}),
  & A_{t,k},A_{t,k+1}\neq\emptyset,\\
  0, & \text{otherwise}.
  \end{cases}
  \label{eq:temporal-pair}
\end{equation}
Temporal fidelity $T_t$ is the mean of these transition scores. With only one
valid observation, we set $T_t=S_t$ rather than granting perfect stability.
The evaluator samples deterministically by timestamp and may add midpoint
samples when match state or edit similarity changes sharply, improving
sensitivity to brief corruption while retaining a bounded frame budget.

Finally, target scores are weighted by normalized target length $w_t=|y_t|$:
\begin{equation}
  S=\frac{\sum_t w_t S_t}{\sum_t w_t},\qquad
  T=\frac{\sum_t w_t T_t}{\sum_t w_t}.
  \label{eq:aggregate}
\end{equation}
The headline metric is their harmonic mean,
\begin{equation}
  \operatorname{TextVScore}=100\cdot\frac{2ST}{S+T},
  \label{eq:textvscore}
\end{equation}
with score zero when either component is zero.

\section{TextVBench-Ref2V}
\label{sec:ref2v}

\subsection{Reference-Conditioned Construction}

\begin{figure}[t]
  \centering
  \includegraphics[width=\linewidth]{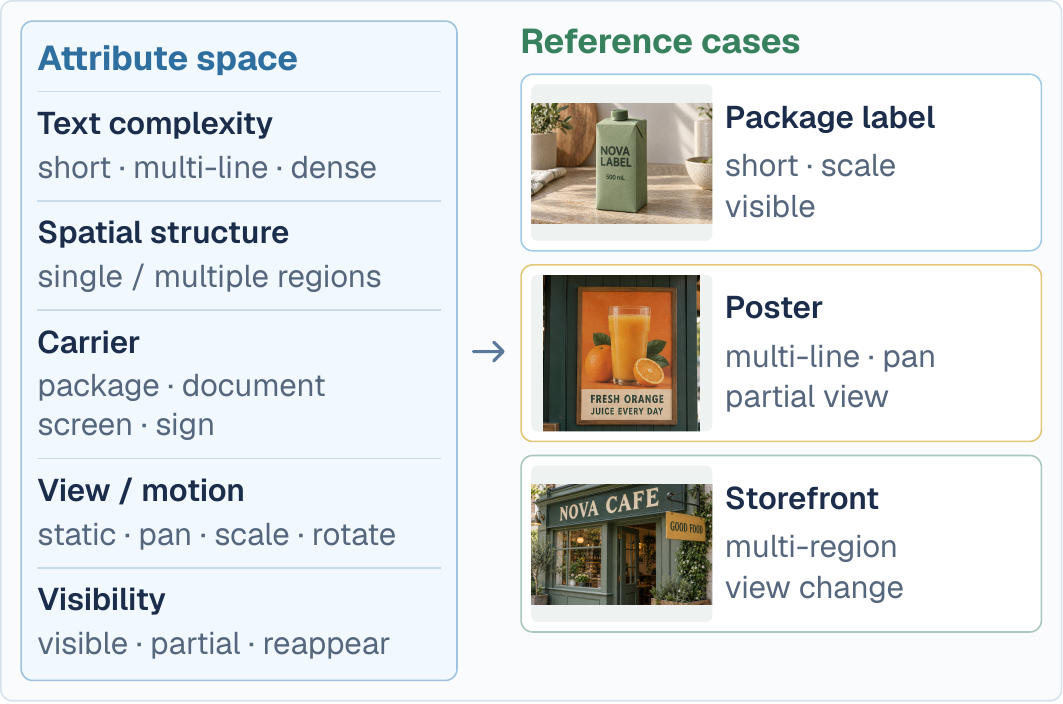}
  \caption{Structured Ref2V construction. Preservation difficulty is controlled
  by combining text complexity, spatial structure, carrier, view/motion, and
  visibility. Generated scenes illustrate three attribute combinations.}
  \label{fig:ref2v-construction}
\end{figure}

Ref2V measures preservation rather than creation. Each example contains a
generation instruction and one or more reference images drawn from diverse
text-bearing carriers, including product packaging, mobile interfaces such as
the AMEX GUI dataset \cite{chai2025amex}, and synthetic application screens.
Source diversity is not itself the benchmark
taxonomy. As illustrated in Fig.~\ref{fig:ref2v-construction}, we organize
construction as a preservation difficulty space
defined by reference text density and length, single- or multi-region spatial
structure, carrier type, and transformations such as scale, camera motion,
device tilt, or object rotation. Single-reference prompts require the carrier
to remain clearly observable and exclude severe occlusion or extreme views;
multi-reference cases use distinct reference IDs to represent view-dependent
activity and reappearance. Prompts request stable preservation of visible text
and layout while avoiding screen transitions or invented content. The released
200-case manifest is source-balanced with 50 AMEX, 50 paper/media, 50 product,
and 50 synthetic-application references; density and preservation results are
reported in Sec.~\ref{sec:experiments}.

\subsection{Reference-to-Video Correspondence}

Targets are discovered at evaluation time. OCR runs on every original
reference image, and each non-empty detection becomes a slot containing a
target string, polygon, and reference ID. All slots from one reference share a
carrier identity.

For a single-reference sample, that reference is activated directly. For a
multi-reference sample, a VLM first routes each generated frame to zero, one,
or several reference IDs using compact reference images labeled with explicit
IDs. Only slots from active references enter target--candidate matching. A
frame with no active reference is recorded but excluded from fidelity and
extra-text aggregation; if no sampled frame activates any reference, all
fidelity scores are zero. Active slots are then matched to generated-frame
OCR candidates using the same marker-image candidate-selection protocol as
T2V. Single-reference samples still use this generated-frame marker image;
they skip only the unnecessary reference-ID routing call. Reference images
provide grounding, but all scored text still comes from OCR candidates. The
reference routing and visibility-aware target trajectories are illustrated in
Fig.~\ref{fig:overview} (steps 02 and 04).

\subsection{Preservation Metrics}

Ref2V uses Eqs.~\ref{eq:spatial-frame}--\ref{eq:textvscore} at slot level.
Within each reference, spatial and temporal scores are weighted by target
length; reference-level scores are then averaged equally so that a reference
with no successful observations cannot disappear from the denominator.
ExtraTextRatio is averaged only over evaluated frames.

Temporal fidelity is computed only across observations for which the target's
reference is active. A single-reference clip keeps its sole reference active
because construction requires that carrier to remain observable; an absent
match is therefore a preservation failure under the requested generation
condition. In multi-reference samples, intervals routed to no reference are
excluded, and matching resumes under the same slot and reference identities
when a reference reappears. This separates expected observability from true
missing text without requiring all reference views in every frame.

\section{Experiments}
\label{sec:experiments}

\subsection{Protocol and Scope}

We evaluate eight text-to-video (T2V) models and six reference-conditioned
video (Ref2V) models using their released or service-recommended inference
settings. Each model is evaluated on the same 300 T2V prompts or 200 Ref2V
references. The evaluator samples at 3~fps with at most 30 frames, uses a
fixed per-sample jitter seed, a 0.80 OCR confidence threshold, at most 20
active candidates per frame, and target batches of at most three. The T2V
fast-match threshold is 0.95. All 2,400 T2V and 1,200 Ref2V videos in this
export completed evaluation successfully. Unless otherwise stated, numbers
are arithmetic means over videos, not means over OCR boxes or frames.
Unresolved candidate selection and Ref2V routing use Qwen3.8-27B with
temperature $0$, top-$p=0.8$, a 512-token response cap, and ID-only JSON
outputs.

\subsection{Overall Comparison}

\begin{table*}[t]
  \centering
  \scriptsize
  \begin{minipage}[t]{0.485\textwidth}
    \centering
    \textbf{T2V: prompt-specified text}\\[2pt]
    \resizebox{\linewidth}{!}{%
    \begin{tabular}{@{}lrrrr@{}}
      \toprule
      Model & Spatial & Temporal & TextVScore & Extra \\
      \midrule
      Wan2.1 T2V-1.3B~\cite{wan2025} & 0.134 & 0.333 & 18.133 & 0.192 \\
      Wan2.2 T2V-A14B~\cite{wan2025} & 0.379 & 0.595 & 43.853 & 0.248 \\
      Wan2.2 TI2V-5B~\cite{wan2025} & 0.293 & 0.475 & 34.577 & 0.146 \\
      MiniMax-H3 & \textbf{0.674} & \textbf{0.701} & \textbf{68.320} & \textbf{0.117} \\
      LongCat-Video~\cite{longcat2025} & 0.086 & 0.299 & 12.598 & 0.319 \\
      Cosmos3-Nano~\cite{nvidia2026cosmos3} & 0.069 & 0.237 & 10.218 & 0.207 \\
      HunyuanVideo-1.5~\cite{hunyuanvideo2025} & 0.090 & 0.122 & 9.792 & 0.206 \\
      Kandinsky-5 Pro~\cite{arkhipkin2025kandinsky50} & 0.229 & 0.392 & 27.359 & 0.281 \\
      \bottomrule
    \end{tabular}}
  \end{minipage}\hfill
  \begin{minipage}[t]{0.485\textwidth}
    \centering
    \textbf{Ref2V: reference text preservation}\\[2pt]
    \resizebox{\linewidth}{!}{%
    \begin{tabular}{@{}lrrrr@{}}
      \toprule
      Model & Spatial & Temporal & TextVScore & Extra \\
      \midrule
      Wan2.2 TI2V-5B~\cite{wan2025} & 0.568 & \textbf{0.566} & 56.388 & \textbf{0.091} \\
      MiniMax-H3 & \textbf{0.576} & 0.557 & 56.465 & 0.104 \\
      LongCat-Video~\cite{longcat2025} & 0.548 & 0.541 & 54.283 & 0.099 \\
      Cosmos3-Nano~\cite{nvidia2026cosmos3} & 0.318 & 0.314 & 31.197 & 0.217 \\
      HunyuanVideo-1.5~\cite{hunyuanvideo2025} & 0.569 & 0.564 & \textbf{56.506} & 0.093 \\
      Kandinsky-5 Pro~\cite{arkhipkin2025kandinsky50} & 0.294 & 0.262 & 27.334 & 0.398 \\
      \bottomrule
    \end{tabular}}
  \end{minipage}
  \caption{Overall results. Spatial and Temporal are in $[0,1]$,
  TextVScore is in $[0,100]$, and Extra is lower-is-better. Bold marks the
  best evaluated value within each track and column. Wan2.2 follows its
  official release's citation to the foundational Wan report.}
  \label{tab:overall-results}
\end{table*}

MiniMax-H3 is clearly strongest on the evaluated T2V comparison: it leads both
fidelity components and has the lowest extra-text ratio. Its TextVScore is
24.47 points above Wan2.2 T2V-A14B, the next model. Ref2V is a
different regime: the top three TextVScores (HunyuanVideo-1.5, MiniMax-H3,
and Wan2.2 TI2V-5B) differ by only 0.119 points. HunyuanVideo-1.5 has the
highest TextVScore, while MiniMax-H3 has the highest spatial fidelity and
Wan2.2 TI2V-5B has the highest temporal fidelity and lowest Extra. We therefore
use track-specific diagnostics below instead of interpreting the small Ref2V
gaps as a reliable total ordering.

\subsection{Scaling With Structured Difficulty}

\begin{figure}[t]
  \centering
  \includegraphics[width=\linewidth]{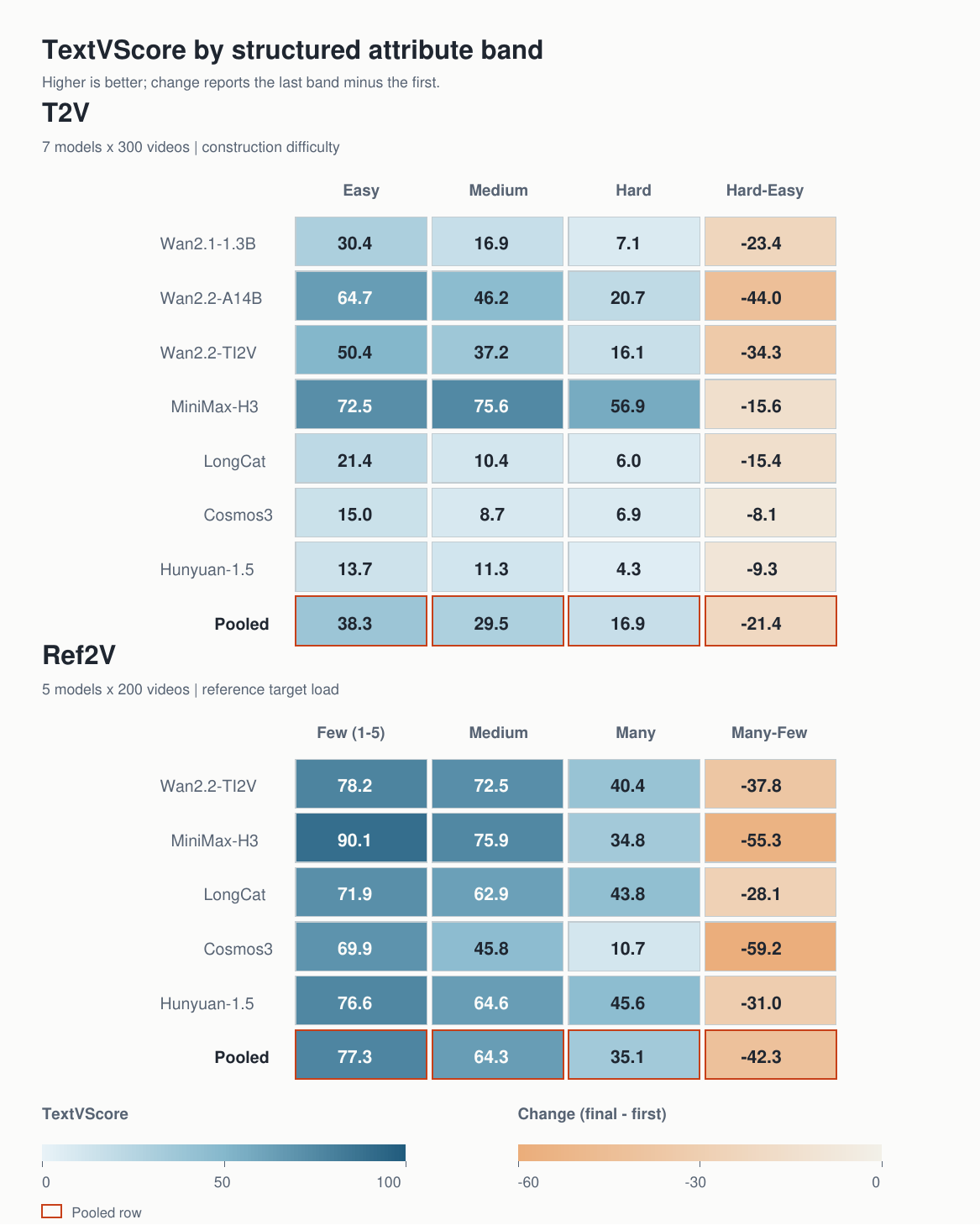}
\caption{Model-wise scaling diagnostics. Each heatmap cell is a video-level
  TextVScore mean; the highlighted pooled row aggregates the detailed
  attribute-slice export.
  The final column reports the last-band minus first-band change, making
  robustness differences explicit. T2V difficulty is fixed by construction,
  while Ref2V load is the number of OCR target slots in the reference. The
  bands are descriptive slices, not matched single-factor experiments.}
  \label{fig:scaling-analysis}
\end{figure}

The benchmark attributes are intended as diagnostic axes, not independent
causal interventions. T2V difficulty jointly increases carrier count, target
count, and text load; Ref2V target load also varies with layout, carrier,
font size, and OCR segmentation. Figure~\ref{fig:scaling-analysis} reports
TextVScore for each model in the detailed attribute-slice export, together with a pooled row and the
change from the first to the last band. This model-by-attribute view makes
robustness differences visible without expanding the main paper with several
cross tables. Kandinsky-5 Pro is included in the overall comparison, while its
current export provides aggregate rows only; it is therefore not included in
the attribute-slice heatmap or its pooled row.

On T2V, pooled TextVScore drops from 38.29 on easy cases to 16.86 on hard
cases, a 21.43-point decrease; Spatial and Temporal drop by 0.207 and 0.216.
ExtraTextRatio also rises from 0.190 to 0.215. The model-level degradation
varies substantially: the hard--easy TextVScore gap is 44.01 points for
Wan2.2 T2V-A14B but 15.56 points for MiniMax-H3, indicating that the
construction levels expose different robustness profiles. The pooled curve
decreases monotonically, whereas individual model curves need not: MiniMax-H3
rises from 72.46 on Easy to 75.60 on Medium before falling to 56.90 on Hard.
We therefore interpret the bands as a stress test rather than a calibrated
difficulty scale.

Ref2V shows a larger common bottleneck. Increasing from few (1--5) to many
(more than 15) target slots lowers pooled TextVScore from 77.34 to 35.06 and
raises ExtraTextRatio from 0.061 to 0.178. Spatial and Temporal fall together,
from approximately 0.775 to 0.360 and 0.347. This pattern is consistent with
crowded references stressing both correspondence and preservation, but the
load bands remain confounded with source, layout, and text length.

\subsection{Failure Profile and Metric Complementarity}

The three headline measurements answer different questions. Table~\ref{tab:failure-profile}
reports pooled failure rates over videos, not additional benchmark scores.
These rates use the detailed attribute-slice export described above and do not
include the Kandinsky-5 Pro aggregate-only row.
T2V has a positive mean Temporal--Spatial gap because a stable OCR observation
can still be the wrong string. Ref2V has nearly matched Spatial and Temporal
means, but half of its videos still have TextVScore below 50.

\begin{table}[t]
  \centering
  \scriptsize
  \setlength{\tabcolsep}{3.5pt}
  \begin{tabular}{@{}lrr@{}}
    \toprule
    Diagnostic & T2V & Ref2V \\
    \midrule
    Spatial failure ($<0.10$) & 49.8\% & 14.2\% \\
    Temporal failure ($<0.10$) & 36.5\% & 16.4\% \\
    TextVScore $=0$ & 29.4\% & 4.9\% \\
    TextVScore $<50$ & 76.4\% & 50.4\% \\
    ExtraTextRatio $>0.20$ & 32.8\% & 19.8\% \\
    Temporal $>$ Spatial & 58.4\% & 26.2\% \\
    Mean Temporal--Spatial & $+0.148$ & $-0.008$ \\
    \bottomrule
  \end{tabular}
  \caption{Pooled failure and complementarity diagnostics. The first two
  rows use the common low-fidelity threshold shown in parentheses;
  ExtraTextRatio captures unassigned readable text and is not folded into
  TextVScore.}
  \label{tab:failure-profile}
\end{table}

The model-level profiles reinforce this separation. HunyuanVideo-1.5 has
Temporal $>$ Spatial in only 24.7\% of T2V videos, with mean $T-S=+0.032$;
its two fidelity components therefore tend to fail together, consistent with
missing targets. In contrast, Wan2.2 T2V-A14B has Temporal $>$ Spatial in
77.7\% of videos and mean $T-S=+0.215$, indicating that OCR observations are
often temporally stable even when the rendered string is wrong. These
contrasting profiles show why Spatial and Temporal must be reported together:
a single frame-level OCR score or stability score would collapse missing text
and stable corruption, while ExtraTextRatio separately exposes unassigned text.

\subsection{Qualitative Evidence}

We use four cases to trace the evaluator's reasoning from a successful target
through increasingly subtle failure modes. Each figure is a single-column
diagnostic panel showing the sampled observations, target-conditioned OCR
crops, spatial scores, and temporal intervals.

\begin{figure}[!b]
  \centering
  \includegraphics[width=0.82\columnwidth]{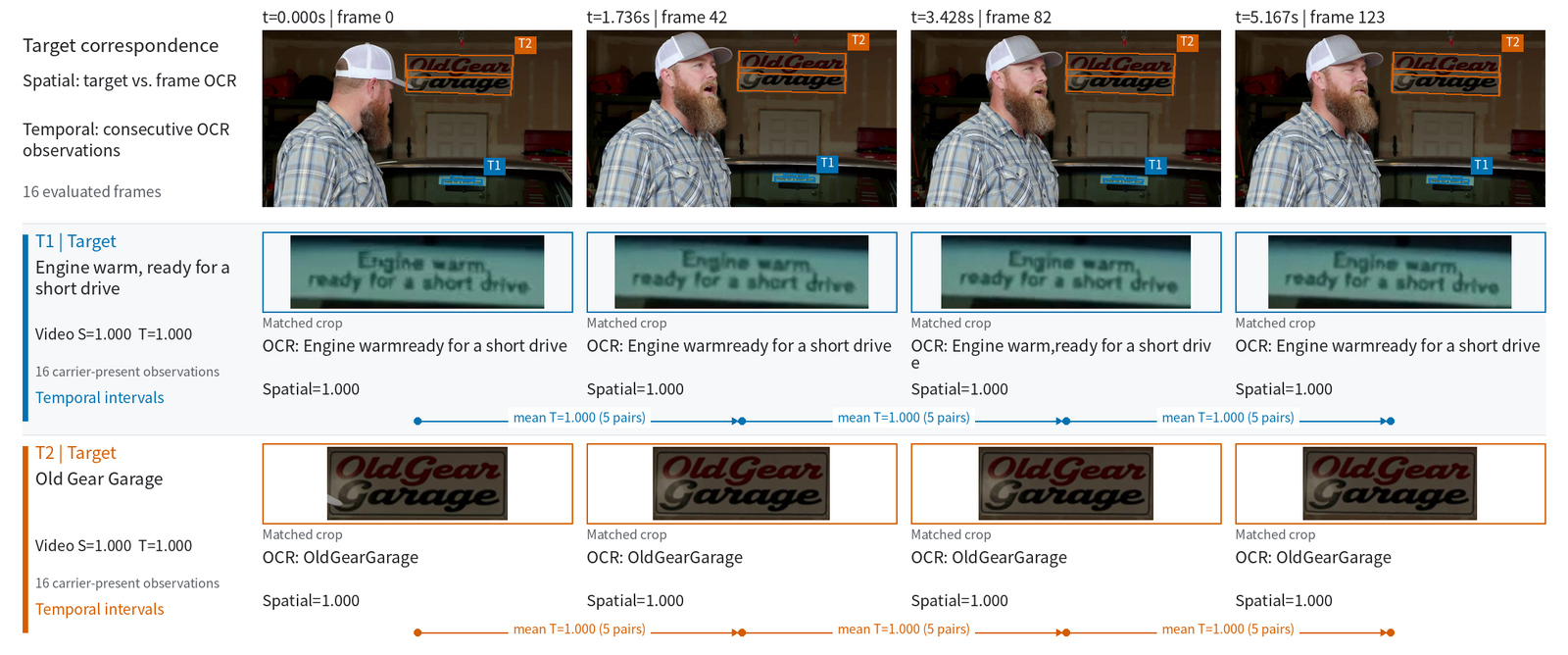}
  \caption{Correct and stable T2V text. Both targets remain matched across
  all sampled observations (Spatial/Temporal $1.000$; TextVScore $100.0$).}
  \label{fig:case-correct-stable}
\end{figure}

\textbf{Correct and stable.} We first establish the desired outcome: both
targets remain correctly matched throughout the video, yielding Spatial $1.000$,
Temporal $1.000$, and TextVScore $100.0$ (Figure~\ref{fig:case-correct-stable}).

\begin{figure}[!ht]
  \centering
  \includegraphics[width=\columnwidth]{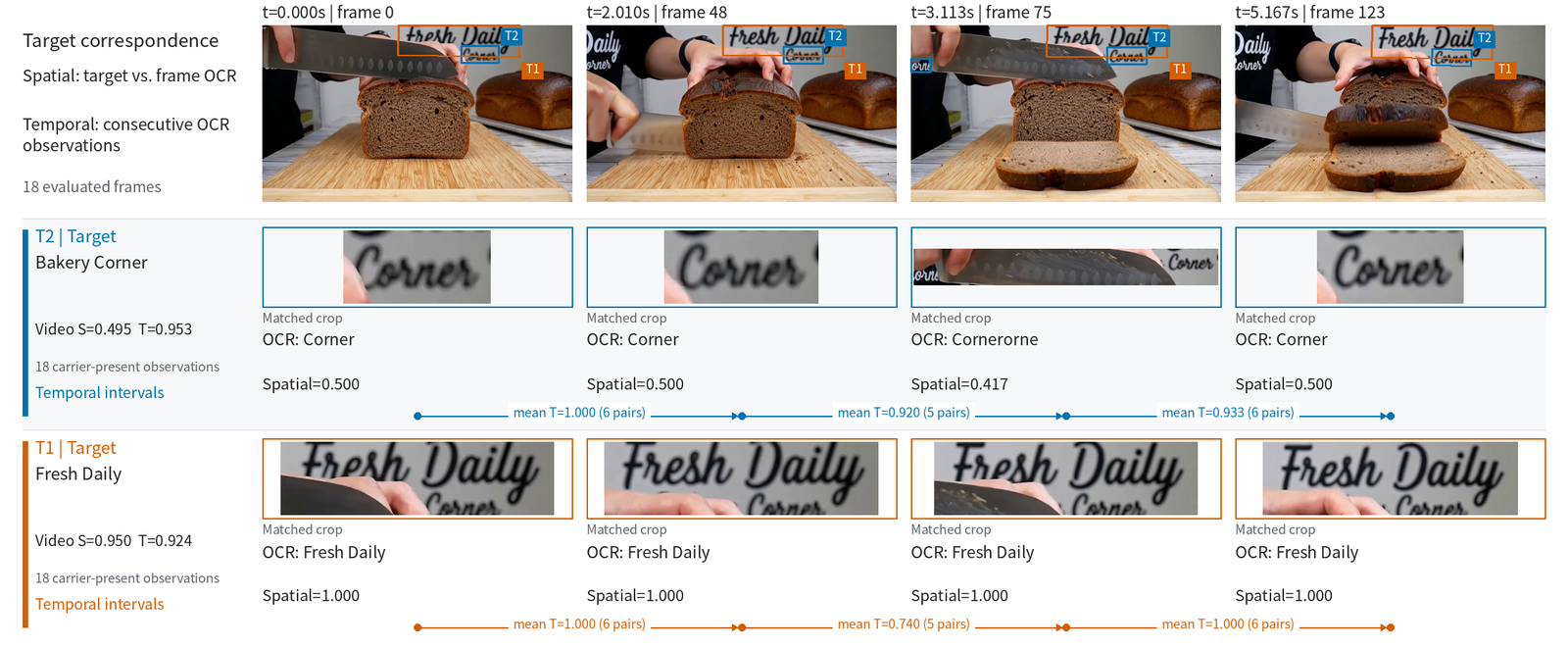}
  \caption{Stable OCR does not imply correct target text. The carrier
  is present throughout, but the requested ``Bakery Corner'' is repeatedly read
  as ``Corner''.}
  \label{fig:case-stable-incorrect}
\end{figure}

\textbf{Stable but incorrect.} The next case isolates textual correctness
from temporal stability. For ``Bakery Corner'', the matched OCR strings
repeatedly omit ``Bakery'', yielding target-level Spatial fidelity of $0.495$
despite Temporal fidelity of $0.953$ (Figure~\ref{fig:case-stable-incorrect}).
Length-weighted aggregation with the second target, ``Fresh Daily'', gives
a video-level Temporal score of $0.940$.

\begin{figure}[!ht]
  \centering
  \includegraphics[width=\columnwidth]{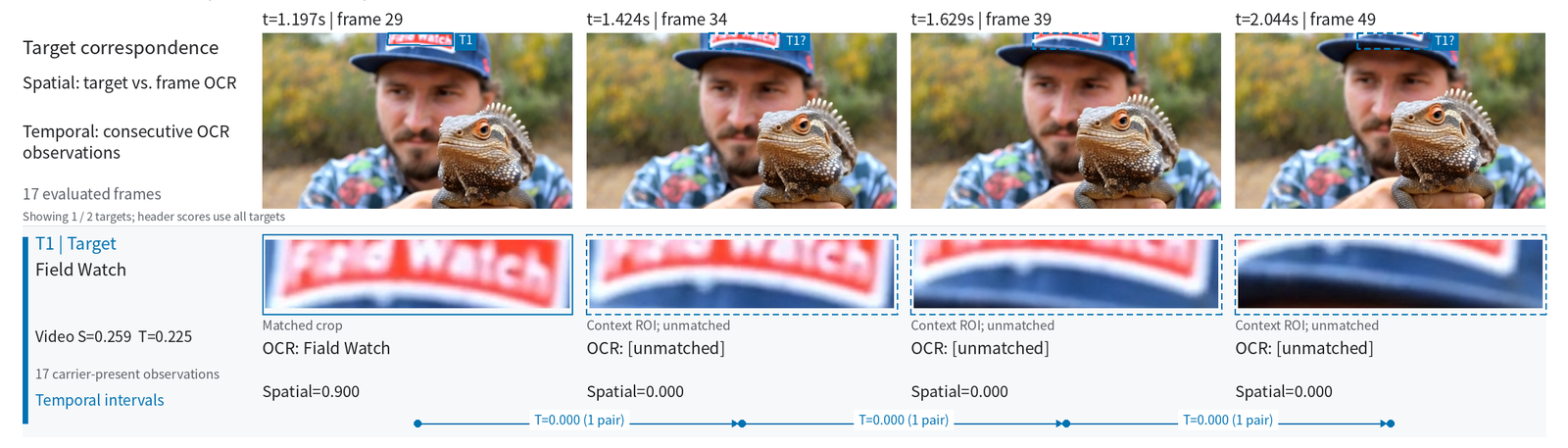}
  \caption{Intermittent correspondence. ``Field Watch'' is matched
  in the first displayed observation and then loses correspondence while the
  carrier remains visible.}
  \label{fig:case-intermittent}
\end{figure}

\textbf{Intermittent correspondence.} A harder T2V failure occurs when
``Field Watch'' is correct briefly and then becomes unmatched even though its
carrier remains visible. Its target-level Spatial and Temporal scores are
$0.259$ and $0.225$; the other equal-length target is completely missing,
which lowers the length-weighted full-video scores to $0.129$ and $0.113$
(Figure~\ref{fig:case-intermittent}).

\textbf{Partial Ref2V preservation.} The final case transfers the same logic
to reference-conditioned generation. Within one reference carrier, the target
``\begin{CJK*}{UTF8}{gbsn}象牙白\end{CJK*}'' is preserved perfectly, whereas ``La meila'' is almost always
unmatched (Spatial $0.0375$, Temporal $0.000$), producing a carrier-level
outcome that would hide target-level variation (Figure~\ref{fig:case-partial-preservation}).

\noindent\begin{minipage}{\columnwidth}
  \centering
  \includegraphics[width=\columnwidth]{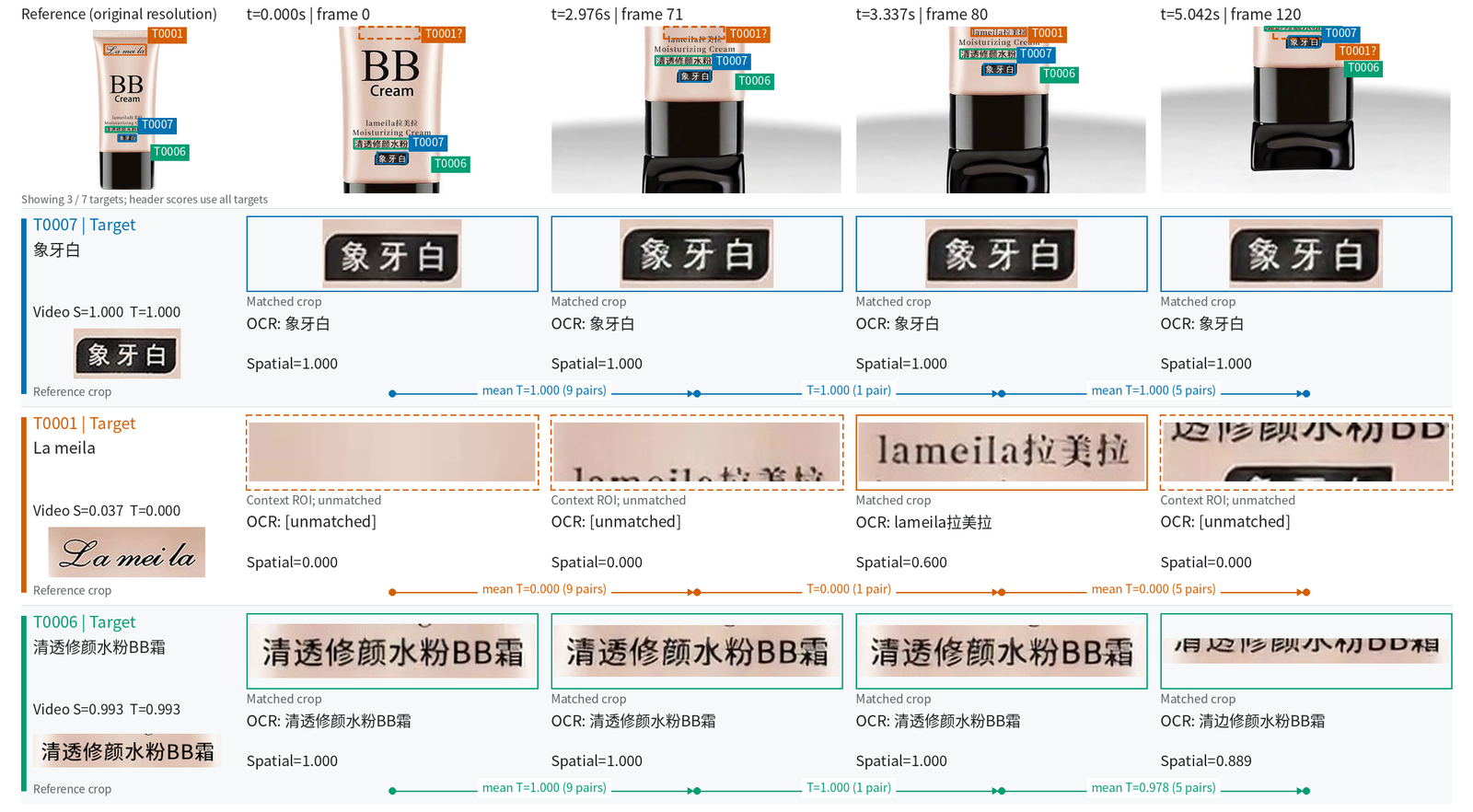}
  \captionof{figure}{Partial Ref2V preservation within one reference carrier.
  Different reference targets receive very different spatial and temporal scores.}
  \label{fig:case-partial-preservation}
\end{minipage}

\section{Limitations}
\label{sec:limitations}

The evaluator depends on OCR and target--candidate correspondence. Tiny or
stylized text, severe blur or perspective, dense layouts, and repeated strings
can therefore confound generation errors with perception or matching errors;
Ref2V also inherits errors from OCR on the reference image. The protocol
further assumes persistent target strings and bounded frame sampling. Dynamic
or state-changing text remains outside the fixed benchmark; multi-reference
Ref2V can handle reference-level inactivity and reappearance, whereas
instance-level T2V re-entry after leaving the view is not modeled by the
current protocol. The reported scores should be interpreted within these
assumptions.

These boundaries motivate TextVBench-Challenge, which we are currently
preparing for a later release as a complementary collection of reproducible
failures beyond the fixed benchmark protocol. The planned release will record
the video or reference, model and prompt, expected text, and observed failure,
covering perceptual extremes, ambiguous correspondence, dynamic text, and
long-range identity changes. It is intended to support specialized evaluators
or human assessment without changing the controlled comparison reported here.

\section{Conclusion}
\label{sec:conclusion}

Visual-text evaluation cannot be reduced to OCR accuracy because OCR outputs
must first be associated with intended semantic targets. Video makes this
association temporal: text identity must remain connected to a carrier or
reference while observations change across frames. \paper makes this
correspondence explicit through complementary T2V creation-and-maintenance and
Ref2V preservation-and-maintenance tracks, followed by target-centric spatial
and temporal scoring.

The structured outputs are useful beyond a leaderboard. Candidate-level
matches and per-target trajectories can support data curation, hard-example
mining, reward modeling, and generate--verify--repair systems.
TextVBench-Challenge extends this direction toward collecting reproducible
failures beyond the current evaluation assumptions (Sec.~\ref{sec:limitations}).
We hope \paper helps shift visual-text
evaluation from frame-wise transcription toward structured spatiotemporal
reasoning.

\FloatBarrier
{
    \small
    \bibliographystyle{ieeenat_fullname}
    \bibliography{main}
}


\end{document}